\documentclass[conference,letterpaper]{IEEEtran}
\ifCLASSINFOpdf
\else
\fi

\usepackage{booktabs}
\usepackage{url}
\usepackage{amssymb}
\usepackage{amsmath}
\usepackage{wrapfig}
\usepackage{enumitem}
\usepackage{mdwlist}
\usepackage{setspace}
\usepackage[caption=false,font=footnotesize]{subfig}
\usepackage{microtype}
\usepackage{color}
\usepackage{algorithm}
\usepackage{tikz}
\usepackage{rotating}
\usepackage{makecell}
\usetikzlibrary{arrows,automata,backgrounds,decorations,fit,petri,positioning,petri,shapes,calc,patterns}
\tikzset{place/.append style={circle,draw=black,thick,inner sep=0pt,minimum size=3mm,label position=below}}
\tikzset{transition/.append style={rectangle,draw=black,thick,inner sep=1.5pt,minimum size=9mm}}
\tikzset{every edge/.append style={->, thin}}
\tikzset{pre/.append style={<-,shorten <=0pt,shorten >=0pt}}
\tikzset{post/.append style={->,shorten >=0pt,shorten <=0pt}}

\newlength{\hatchspread}
\newlength{\hatchthickness}
\newlength{\hatchshift}
\newcommand{\hatchcolor}{}
\tikzset{hatchspread/.code={\setlength{\hatchspread}{#1}},
	hatchthickness/.code={\setlength{\hatchthickness}{#1}},
	hatchshift/.code={\setlength{\hatchshift}{#1}},
	hatchcolor/.code={\renewcommand{\hatchcolor}{#1}}}
\tikzset{hatchspread=3pt,
	hatchthickness=0.4pt,
	hatchshift=0pt,
	hatchcolor=black}
\pgfdeclarepatternformonly[\hatchspread,\hatchthickness,\hatchshift,\hatchcolor]
{custom north west lines}
{\pgfqpoint{\dimexpr-2\hatchthickness}{\dimexpr-2\hatchthickness}}
{\pgfqpoint{\dimexpr\hatchspread+2\hatchthickness}{\dimexpr\hatchspread+2\hatchthickness}}
{\pgfqpoint{\dimexpr\hatchspread}{\dimexpr\hatchspread}}
{
	\pgfsetlinewidth{\hatchthickness}
	\pgfpathmoveto{\pgfqpoint{0pt}{\dimexpr\hatchspread+\hatchshift}}
	\pgfpathlineto{\pgfqpoint{\dimexpr\hatchspread+0.15pt+\hatchshift}{-0.15pt}}
	\ifdim \hatchshift > 0pt
	\pgfpathmoveto{\pgfqpoint{0pt}{\hatchshift}}
	\pgfpathlineto{\pgfqpoint{\dimexpr0.15pt+\hatchshift}{-0.15pt}}
	\fi
	\pgfsetstrokecolor{\hatchcolor}
	\pgfusepath{stroke}
}

\newtheorem{definition}{Definition}
\newcommand{\fleche}{\longrightarrow}
\newcommand{\flsup}[1]{\stackrel{#1}{\fleche}}
\newcommand{\step}[1]{\flsup{#1}}           
\newcommand{\Lan}                 {\mathfrak{L}}

\newcommand{\pre}[1]{\bullet #1}
\newcommand{\APN}{\mathit{APN}}
\newcommand{\LN}{\mathit{LN}}
\newcommand{\MF}{\mathit{MF}}

\def\tproj{\!\!\upharpoonright}

\begin{document}
%
\title{Heuristic Approaches for Generating Local Process Models through Log Projections}

\author{\IEEEauthorblockN{Niek Tax, Natalia Sidorova, Wil M. P. van der Aalst}
\IEEEauthorblockA{Department of Mathematics and Computer Science\\
Eindhoven University of Technology\\
P.O. Box 513, 5600MB Eindhoven, The Netherlands\\
Email: \{n.tax,n.sidorova,w.m.p.v.d.aalst\}@tue.nl}
\and
\IEEEauthorblockN{Reinder Haakma}
\IEEEauthorblockA{Philips Research\\
	Prof. Holstlaan 4, 5665 AA Eindhoven\\
	The Netherlands\\
	Email: reinder.haakma@philips.com}
}


%


\maketitle

\begin{abstract}
Local Process Model (LPM) discovery is focused on the mining of a set of process models where each model describes the behavior represented in the event log only partially, i.e. subsets of possible events are taken into account to create so-called local process models. Often such smaller models provide valuable insights into the behavior of the process, especially when no adequate and comprehensible single overall process model exists that is able to describe the traces of the process from start to end. The practical application of LPM discovery is however hindered by computational issues in the case of logs with many activities (problems may already occur when there are more than 17 unique activities). In this paper, we explore three heuristics to discover subsets of activities that lead to useful log projections with the goal of speeding up LPM discovery considerably while still finding high-quality LPMs. We found that a Markov clustering approach to create projection sets results in the largest improvement of execution time, with discovered LPMs still being better than with the use of randomly generated activity sets of the same size. Another heuristic, based on log entropy, yields a more moderate speedup, but enables the discovery of higher quality LPMs. The third heuristic, based on the relative information gain, shows unstable performance: for some data sets the speedup and LPM quality are higher than with the log entropy based method, while for other data sets there is no speedup at all.\looseness=-1
\end{abstract}


%
\IEEEpeerreviewmaketitle

\section{Introduction}
Process mining is an area of research that combines methods and techniques from computational intelligence, data mining, and process analysis to extract novel insights from event data \cite{Aalst2016}. One of the most prominent tasks within the process mining field is \emph{process discovery}, where the goal is to discover a process model that accurately describes some sequences -- called \emph{traces} -- of event data from start to end. In some application domains a high degree of variability can be found in such traces. An example of such a domain is human behavior \cite{Tax2016b}, with events registered e.g. in smart homes, by wearables, or manually in so called LifeLogs. Another application domain, where the variability in the process is often large is the medical workflows of patient care \cite{Mans2008}.

Existing process discovery algorithms (e.g. \cite{Leemans2013,Werf2008}) often fail to generate insightful models on such event logs and generate process models in which any sequence of events is allowed, often referred to as the \emph{flower model}. More insight in such event logs can often be obtained using declarative process discovery algorithms (e.g. \cite{Maggi2011,Schonig2015}). Declarative models describe the behavior through a set of constraints on activities (event types), e.g. binary constraint ``each activity $a$ is always followed by activity $b$''. They primarily focus on binary constrains, with a limited set of non-binary extensions by branching, like ``each activity $a$ is always followed by activity $b$ or activity $c$''.
Richer models of unstructured processes can be discovered using \emph{Local Process Model (LPM)} discovery \cite{Tax2016}. LPMs describe frequent behavioural patterns in a process modeling notation (e.g. Petri nets, BPMN, UML activity diagrams), allowing each pattern to have full expressive power of the respective process model notation. LPMs allow for complex relations between activities.

\begin{figure}
\centering
\scalebox{0.735}{
	\begin{tikzpicture}
	[node distance=1.4cm,
	on grid,>=stealth',
	bend angle=20,
	auto,
	every place/.style= {minimum size=0.1mm},
	every transition/.style = {minimum size = 10mm, text width=1.8cm,align=center},
	transitionH/.style={rectangle, thick, fill=black, minimum width=3mm, inner ysep=9pt }
	]
	\node [place, tokens = 1] (p){};
	\node [transition] (2) [right = of p,font=\scriptsize] {\textbf{CONTINUOUS INVASIVE MECH} 3638/8291}
	edge [pre] node[auto] {} (p);
	\node [place] (p3) [right = of 2] {}
	edge[pre] node[auto] {} (2);
	\node [transition] (t1) [above right = of p3,font=\scriptsize]{\textbf{ARTERIAL CATHETERIZATION} 1028/2478}
	edge[pre] node[auto] {} (p3);
	\node [transitionH] (t3) [below right = of p3,fill=black] {}
	edge[pre] node[auto] {} (p3);
	\node [place] (p8) [right=of t3]{}
	edge[pre] node[auto] {}(t3);
	\node [transitionH] (t4) [right = of p8,fill=black]{}
	edge[post] node[auto] {}(p8);
	\node [place] (p9) [above right = of t4]{}
	edge[post, bend left] node[auto] {}(t4);
	\node [transition] (t6) [left = of p9,font=\scriptsize]{\textbf{EXT INFUS CONC NUTRITION} 2710/5543}
	edge[post] node[auto] {}(p9)
	edge[pre] node[auto] {}(p8);
	\node [transitionH] (t5) [above left = of p9,fill=black]{}
	edge[pre, bend left] node[auto] {}(p9);
	\node [place,pattern=custom north west lines,hatchspread=1.5pt,hatchthickness=0.25pt,hatchcolor=gray] (p7) [right=of t1] {}
	edge[pre] node[auto] {}(t1)
	edge[pre] node[auto] {}(t5);
	\end{tikzpicture}}

	\vspace{0.35cm}
	
	\scalebox{0.735}{
	\begin{tikzpicture}
	[node distance=1.4cm,
	on grid,>=stealth',
	bend angle=20,
	auto,
	every place/.style= {minimum size=1mm},
	every transition/.style = {minimum size = 10mm, text width=1.8cm,align=center},
	transitionH/.style={rectangle, thick, fill=black, minimum width=3mm, inner ysep=9pt }
	]
	\node [place, tokens = 1] (p){};
	\node [transitionH] (2) [right = of p,fill=black] {}
	edge [pre] node[auto] {} (p);
	\node [place] (p3) [above right = of 2] {}
	edge[pre] node[auto] {} (2);
	\node [place] (p4) [below right = of 2] {}
	edge[pre] node[auto] {} (2);
	\node [transition] (t1) [right = of p3,font=\scriptsize]{\textbf{CONTINUOUS INVASIVE MECH} 6080/8291}
	edge[pre] node[auto] {} (p3);
	\node [transition] (t3) [right = of p4,font=\scriptsize] {\textbf{INSERT ENDOTRACHEAL TUBE} 6080/6432}
	edge[pre] node[auto] {} (p4);
	\node [place] (p8) [right=of t3]{}
	edge[pre] node[auto] {}(t3);
	\node [place] (p9) [right = of t1]{}
	edge[pre] node[auto] {}(t1);
	\node [transitionH] (t6) [below right = of p9,fill=black]{}
	edge[pre] node[auto] {}(p9)
	edge[post] node[auto] {}(p8);
	\node [place,pattern=custom north west lines,hatchspread=1.5pt,hatchthickness=0.25pt,hatchcolor=gray] (p7) [right=of t6] {}
	edge[pre] node[auto] {}(t6);
	\end{tikzpicture}
	}
	
	\vspace{0.35cm}
	
	\scalebox{0.735}{
		\begin{tikzpicture}
		[node distance=1.4cm,
		on grid,>=stealth',
		bend angle=20,
		auto,
		every place/.style= {minimum size=1mm},
		every transition/.style = {minimum size = 7mm, text width=1.8cm,align=center},
		transitionH/.style={rectangle, thick, fill=black, minimum width=3mm, inner ysep=9pt }
		]
		\node [place, tokens = 1] (p){};
		\node [transitionH] (2) [below right = of p,fill=black] {}
		edge [post] node[auto] {} (p);
		\node [transition] (3) [above right = of p,font=\scriptsize] {{\textbf{HEMODIA- LYSIS} 1314/1314}}
		edge [pre] node[auto] {} (p);
		\node [place] (p3) [above right = of 2] {}
		edge[post] node[auto] {} (2)
		edge[pre] node[auto] {} (3);
		\node [transition] (t1) [right = of p3,font=\scriptsize]{\textbf{VENOUS CATH FOR RENAL DI} 339/700}
		edge[pre] node[auto] {} (p3);
		\node [place,pattern=custom north west lines,hatchspread=1.5pt,hatchthickness=0.25pt,hatchcolor=gray] (p9) [right = of t1]{}
		edge[pre] node[auto] {}(t1);
		\end{tikzpicture}	
	}
	
	\vspace{0.35cm}
	
	\scalebox{0.735}{
		\begin{tikzpicture}
		[node distance=1.4cm,
		on grid,>=stealth',
		bend angle=20,
		auto,
		every place/.style= {minimum size=1mm},
		every transition/.style = {minimum size = 7mm, text width=1.8cm,align=center},
		transitionH/.style={rectangle, thick, fill=black, minimum width=3mm, inner ysep=9pt }
		]
		\node [place, tokens = 1] (p){};
		\node [transitionH] (2) [below right = of p,fill=black] {}
		edge [post] node[auto] {} (p);
		\node [transition] (3) [above right = of p,font=\scriptsize] {{\textbf{CONTINUOUS INVASIVE MECH} 8291/8291}}
		edge [pre] node[auto] {} (p);
		\node [place] (p3) [above right = of 2] {}
		edge[post] node[auto] {} (2)
		edge[pre] node[auto] {} (3);
		\node [transition] (t1) [right = of p3,font=\scriptsize]{\textbf{NON-INVASIVE MECHANICAL} 790/2193}
		edge[pre] node[auto] {} (p3);
		\node [place,pattern=custom north west lines,hatchspread=1.5pt,hatchthickness=0.25pt,hatchcolor=gray] (p9) [right = of t1]{}
		edge[pre] node[auto] {}(t1);
		\end{tikzpicture}	
	}
	\caption{Local process models discovered using projection set discovery on the MIMIC II medical procedure data set}
	\label{fig:motivating_example}
	\vspace{-0.3cm}
\end{figure}
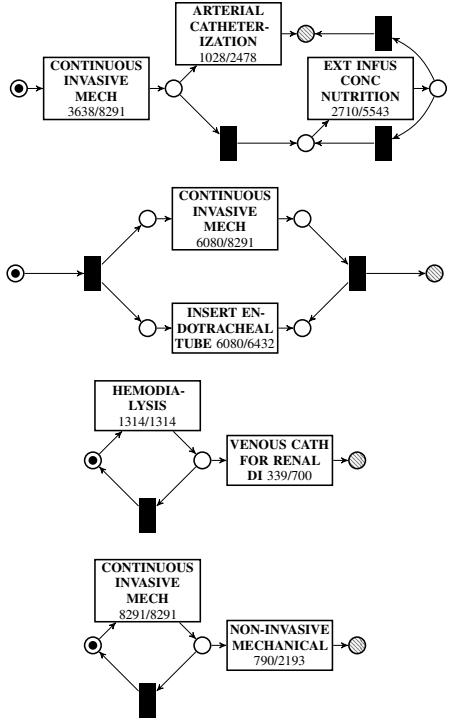

Fig. \ref{fig:motivating_example} shows some of the LPMs discovered on the log extracted from MIMIC-II \cite{Saeed2002}, a medical database containing $147461$ logged medical procedure events from $1734$ activities for $28280$ patients collected between 2001 and 2008 from multiple intensive care units (ICUs) in the USA. The first LPM indicates that the placement of continuous invasive mechanical ventilation on a patient is $3638$ out of $8291$ times followed by  either arterial catheterization or by one or more instances of infusion of concentrated nutrition. The second LPM shows that the placement of invasive mechanical ventilation frequently co-occurs with the insertion of an endotracheal tube. The third LPM shows that a sequence of one or more hemodialysis events is always followed by the placement of a venous catheter for renal dialysis, and from the numbers we can deduce that there are on average almost $4$ hemodialysis events before one placement of a venous catheter. The fourth LPM in Fig. \ref{fig:motivating_example} shows that all placements of a continuous invasive mechanical ventilation are eventually followed by the placement of a non-invasive mechanical ventilation. However, on average $8$ continuous invasive mechanical ventilations have been applied before the placement of a non-invasive mechanical ventilation.

The LPMs in Fig. \ref{fig:motivating_example} cannot be discovered using the brute-force approach described in \cite{Tax2016}. There are 1734 activities in the log, yielding an incredible number of models. The computational complexity of LPM discovery grows combinatorially with the number of activities in the event log, hindering the practical application to many real life event logs. However, LPMs can be discovered on a projection on a subset of the activities in the log, as long as the set of activities projected on contains at least the activities in the LPM. The top LPM in Fig. \ref{fig:motivating_example} can for example be discovered using a projection on just the activities \textit{CONTINUOUS INVASIVE MECH}, \textit{ARTERIAL CATHETERIZATION}, and \textit{EXT INFUS CONC NUTRITION}. In this paper, we explore and compare three heuristic approaches to LPM mining, using intelligently chosen subsets of the activities. By applying LPM discovery to an event log projected on the subsets of activities that are discovered with the heuristic approaches the computational time of discovery decreases considerably. For example, the LPMs in Fig. \ref{fig:motivating_example} were discovered in $12$ seconds using the Markov clustering approach that we will discuss in Section \ref{sec:approach}.\looseness=-1

We start by introducing basic concepts and notations in Section \ref{sec:preliminaries}. In Section \ref{sec:lpm} we discuss quality criteria and rankings of Local Process Models. In Section \ref{sec:approach} we introduce the three heuristic approaches to discovery of projection sets. We introduce an experimental setup for evaluation of heuristics and describe five real-life event logs used for evaluation in Section \ref{sec:evaluation_methodology}. Section \ref{sec:results} contains the results of the experiments and their discussion. We discuss related work in Section \ref{sec:related} and conclude this paper in Section \ref{sec:conclusion}.

\section{Preliminaries}
\label{sec:preliminaries}
In this section we introduce concepts used in later sections of this paper.

$X^*$ denotes the set of all sequences over a set $X$ and $\sigma=\langle a_1,a_2,\dots,a_n\rangle$ a sequence of length $n$; $\langle\rangle$ is the empty sequence and $\sigma_1 \cdot \sigma_2$ is the concatenation of sequences $\sigma_1,\sigma_2$.

In the context of process logs, we assume the set of all \emph{process activities} $\Sigma_L$ to be given. An \emph{event} $e$ in an event log is the occurrence of an activity $e\!\in \!\Sigma_L$. We call a sequence of events $\sigma\in {\Sigma^*_L}$ a \emph{trace}. An \emph{event log} $L\!\in\! \mathbb{N}^{{\Sigma^*_L}}$ is a finite multiset of traces. For example, the event log $L=[\langle a,b,c\rangle^2,\langle b,a,c\rangle^3]$ consists of 2 occurrences of trace $\langle a,b,c\rangle$ and three occurrences of trace $\langle b,a,c\rangle$.
A projection of sequence $\sigma\!\!\in\!\!X^*$ on a subset $X'\!\!\subseteq\!\!X$ is denoted by $\sigma\!\tproj_{X'}$. For example, $\langle a,b,c,a,b,c\rangle\!\tproj_{\{a,c\}}\!=\!\!\langle a,c,a,c \rangle$. We also lift the projection operator to multi-sets of sequences. For example, $[\langle a,b,c,a,b,c\rangle^3,\langle a,c,a,d,c\rangle^2,\langle a,c,d,c\rangle^4]\!\!\tproj_{\{a,c\}}=
[\langle a,c,a,c\rangle^5,\langle a,c,c \rangle^4]$.\looseness=-1

In the context of process mining, commonly used statistics over the log are related to the activities directly following/preceding each other in the traces of the log \cite{Aalst2016}. The \emph{directly follows ratio} $\textit{dfr}(a,b,L)$ for activities $(a,b)$ in log $L$ is the ratio of occurrences of $a$ that are directly followed by $b$ in event log $L$ to the total number of the occurrences of $a$ in $L$. The \emph{directly precedes ratio} of $(a,b)$ in log $L$, denoted $\textit{dpr}(a,b,L)$, is the ratio of occurrences of $a$ directly preceded by a $b$ to the total number of the occurrences of $a$ in $L$. Assuming an arbitrary but fixed order over the activities of $\Sigma_L$, $\textit{dpr}(a,L)$ and $\textit{dfr}(a,L)$ represent respectively the vectors of values $\textit{dpr}(a,b,L)$, $\textit{dfr}(a,b,L)$ for all $b\in \Sigma_L$. 


%

Petri nets is a process modeling formalism frequently used in process modeling and process mining. A Petri net is a directed bipartite graph consisting of places (depicted as circles) and transitions (depicted as rectangles), connected by arcs. Transitions represent activities, while places represent the enabling conditions of transitions. Labels are assigned to transitions to indicate the type of activity that they model. A special label $\tau$ is used to represent invisible transitions (depicted as black rectangles), which are only used for routing purposes and not recorded in the execution log.
\begin{definition}[Labeled Petri net]
	\label{def:lpn}
	A \emph{labeled Petri net} $N=\langle P,T,F,\Sigma_M,\ell\rangle$ is a tuple where $P$ is a finite set of places, $T$ is a finite set of transitions such that $P \!\cap\!T\!=\!\emptyset$,  $F\subseteq(P \times T) \cup (T \times P)$ is a set of directed arcs, called the flow relation, $\Sigma_M$ is a finite set of labels representing activities, with $\tau \notin \Sigma_M$ being a label representing  invisible events, and $\ell:T\rightarrow \Sigma_M\cup \{\tau\}$ is a labeling function that assigns a label to each transition.
\end{definition}
For a node $n \in (P \cup T)$ we use $\bullet n$ and $n \bullet$ to denote the set of input and output nodes of $n$. A state of a Petri net is defined by its \emph{marking} $M \in \mathbb{N}^{P}$ being a multiset of places. A marking is graphically denoted by putting $M(p)$ tokens on each place $p\in P$. A pair $(N,M)$ is called a marked Petri net. State changes occur through transition firings. A transition $t$ is enabled (can fire) in a given marking $M$ if each input place $p\in \pre{t}$ contains at least one token. Once a transition fires, one token is removed from each input place of $t$  and one token is added to each output place of $t$, leading to a new marking.
A firing of a transition $t$ leading from marking $M$ to marking $M'$ is denoted as $M \step{t} M'$. $M_1 \step{\gamma} M_2$ indicates that $M_2$ can be reached from $M_1$ through firing sequence $\gamma\in T^*$.\looseness=-1

Often, it is useful to consider a Petri net in combination with an initial marking and a set of possible final markings. This allows us to define the language accepted by the Petri net as a set of finite sequences of activities.

\begin{definition}[Accepting Petri Net]
	An \emph{accepting Petri net} is a triple $\APN = (N,M_0,\MF)$, where $N$ is a labeled Petri net, $M_0\in\mathbb{N}^p$ is its initial marking, and $\MF\subset\mathbb{N}^p$ is its set of possible final markings, such that $\forall_{M_1,M_2 \in \MF} \;M_1\!\not\subset\! M_2$. A sequence $\sigma\in \Sigma_M^*$ is a \emph{trace} of an accepting Petri net $\APN$ if there exists a firing sequence $M_0\step{\gamma} M_f$ such that $M_f\!\in\!\MF$, $\gamma\in T^*$ and $\ell(\gamma)=\sigma$. The \emph{language} $\Lan(APN)$ is the set of all its traces, which can be infinite when $\APN$ contains one or more loops.\looseness=-1
\end{definition}

The four models in Fig. \ref{fig:motivating_example} are  accepting Petri nets. Places that belong to the initial marking contain a token and places belonging to a final marking are simply marked as $\begin{tikzpicture}
[node distance=1.4cm,
on grid,>=stealth',
bend angle=20,
auto,
every place/.style= {minimum size=0.1mm},
]
\node [place,pattern=custom north west lines,hatchspread=1.5pt,hatchthickness=0.25pt,hatchcolor=gray] {};
\end{tikzpicture}$, since the nets in the figure have a single final marking. The language of the accepting Petri net at the bottom consists of all the sequences starting with one of more $\textit{CONTINUOUS INVASIVE MECH}$ followed by $\textit{NON-INVASIVE MECHANICAL}$.\\

\section{Local Process Models and Their Rankings}
\label{sec:lpm}
Local Process Model (LPM) discovery generates a set of LPMs that each individually describe some frequent behavior in the event log. The quality of an LPM $\LN$ is calculated using segmentation of traces from a log into sequences $\gamma_0\xi_1\gamma_1\xi_2\ldots\xi_k\gamma_k$, with $\xi_i\!\in\!\Lan(\LN)$ and $\gamma_i\!\notin\!\Lan(\LN)$, such that the number of events in $\xi_1\ldots\xi_k$ is maximized: the higher the number of events in $\xi$ segments, the larger the share of traces of the log explained by the LPM. The ranking of Local Process Models is based on a weighted average over five quality criteria in a zero to one range, as described in \cite{Tax2016}:\looseness=-1
\begin{description}
	\item[Support] {The frequency of the behavior described by the LPM in the event log, i.e. the number of trace segments in the log that fit $\Lan(\LN)$.}
	\item[Confidence] {The share of events of the activities described by the LPM that abide to the behavior described by the LPM, i.e., are in a $\xi_i\!\in\!\Lan(\LN)$ segment.}
	\item[Language fit]{The ratio of traces from $\Lan(\LN)$ that were observed at least once in the event log to $|\Lan(\LN)|$ (bounded up to a certain length for models with loops).}
	\item[Determinism]{This metric reflects the average number of enabled transitions in each marking of the LPM reached while replaying the event log on the LPM. The intuition behind this is that a model with a higher degree of non-determinism captures less information about the ordering of events.}
	\item[Coverage]{The frequency of the activities described in the LPM in the event log.}
\end{description}
In \cite{Tax2016} we developed an incremental procedure for building LPMs, starting from models with two activities, and recursively extending them to more activities.
The support and the determinism quality dimensions can be used there for pruning thanks to their monotonicity with respect to model extensions, resulting in speedup of LPM discovery because of a smaller search space of LPMs.  

\section{Discovering Log Projection Sets}
\label{sec:approach}
Local Process Models (LPMs) only contain a subset of the activities of the log and each LPM can in principle be discovered on any projection of the log containing the activities used in this LPM. In this section we describe three heuristics for discovery of \emph{projection sets} -- subsets of activities to be used for projecting the log. Each heuristic described takes an event log as input and produces a set of projection sets. These projection sets could potentially be overlapping, which is a desired property as interesting patterns might exist within a set of activities $\{A,B,C,D\}$, as well as within a set of activities $\{A,B,C,E\}$, and discovering on both subsets individually is faster than discovering once on $\{A,B,C,D,E\}$. None of the projection sets in this set is a subset of another projection set to avoid double work, as each LPM that can be discovered on a certain projection set can also be discovered on a superset of that same projection set.

\subsection{An approach based on Markov clustering}
Markov clustering \cite{Dongen2008,Dongen2000} is a fast and scalable clustering algorithm for graphs that is based on simulation of flow in graphs. The main intuition behind Markov clustering is that, while performing a random walk on a graph, the likelihood of transitioning between two members of the same cluster is higher than the likelihood of transitioning between two nodes that are in different clusters. Markov clustering takes as input a Markov matrix, i.e. a matrix that describes the transition probabilities in a Markov chain.
We generate a matrix $M$ that represents the \emph{connectedness} of two activities by using the directly-follows and directly-precedes ratios: $M_{i,j} = \sqrt{\textit{dpr}(i,j,L)^2+ \textit{dfr}(j,i,L)^2}$, using the $L_2$ norm of the directly precedes ratio and the directly follows ratio. A Markov matrix $M'$ is obtained by normalizing $M$ row-wise. The intuition behind using both $\textit{dpr}$ and $\textit{dfr}$ combined instead of either of the two is that it can be both of importance that activity $i$ is often followed by $j$ and that $j$ is often preceded by $i$; if either of the two is true than there is apparently some relation $i\rightarrow j$. Applying Markov clustering to $M'$ results in a set of clusters of activities, where we use each activity cluster as a projection set. Markov clustering simulates a random walk over a graph by alternating \emph{expansion}, taking the power of this matrix, and \emph{inflation}, taking the entrywise power of this matrix. The clusters generated by Markov clustering can overlap, which is a desired property in projection set discovery. The inflation parameter of the Markov clustering algorithm is known to be the main parameter in determining the granularity of the clustering obtained \cite{Dongen2000} with Markov clustering.

\subsection{Log entropy based approach}
Another approach to generate projections sets is to form groups of activities such that the categorical probability distributions over activities preceding or following an activity in the projected logs are peaked, i.e. far away from the discrete uniform distribution. When after an occurrence of activity $a$ observing each other type of activity as the next event in the projected log is equally likely, the activities can be considered to be independent of each other. If, on the other hand, the occurrence of activity $a$ conveys information on the next event and e.g. makes it very likely that some other activity $b$ is going to be observed next, the activities in the projection set are likely to be somehow related.

We use the standard entropy function over categorical probability distribution $X$ over $|X|$ elements: $\relpenalty10000\binoppenalty10000 H(X)=\sum_{x\in X}-x\cdot log_2(x)$. We calculate the total entropy $\textit{Ent}(L)$ of the log statistics in original event log $L$ as:
$$\textit{Ent}(L) = \sum_{a\in{\Sigma_L}}(\mathit{H}(\textit{dfr}(a,L))+\mathit{H}(\textit{dpr}(a,L))).$$

We choose an entropy ratio threshold parameter $r$ to indicate the maximum entropy of the log statistics of a log projection.
We start from a set of elementary log projection sets, $S_{1}=\{\{a\}|a\in\Sigma_L\}$. We proceed iteratively as follows: we define
$S_{i+1}=\{A\cup B|A\in S'_i,B\in S_{1}:B\nsubseteq A\}$ and select those projection sets of $S_{i+1}$ that lead to the log projections whose total entropy does not exceed the threshold defined by parameter $r$: $S'_{i+1}=\{A|A\in S_{i+1} \wedge \textit{Ent}(L\tproj_{A})\le r\cdot \textit{Ent}(L)\}$.
The procedure stops if $S'_{i+1}=\emptyset$ or $S_{i+1}=\Sigma_L$.
Since LPMs that can be discovered using some projection set can also be discovered using a superset of this projection set, projection sets that are subsets of other projection sets are removed from the final result: $S_{\textit{final}} = \{A|\exists S'_i: (A\in S'_i \wedge (\forall S'_j ,\forall B\in S'_j: A\not\subset B))\}$.\looseness=-1

\subsection{Maximal Relative Information Gain based approach}
A more local perspective on entropy-based projection set discovery would be to compare a projection set with the projection set of the previous time step, instead of the original log. We add an activity to a projection set when adding it strongly decreases the entropy of at least one of the categorical probability distributions over following or preceding activities, even when the entropy of other categorical probability distributions might increase. The intuition behind this is that any decrease in entropy of a categorical probability distribution over following or preceding activities indicates that some pattern is getting stronger by adding that activity and that a LPM could potentially be found.

We define the Maximal Relative Information Gain (MRIG) of projection set $A$ over projection set $B$, with $A\supset B$, on event log $L$, $\textit{MRIG}(A,B,L)$, as the maximal relative information gain over all the log statistics on $L$, that is the ratio of bits for encoding the log statistic that is most decreased by growing the projection set:\looseness=-1

\noindent$\mathit{MRIG}(A,B,L) = \max_{a\in B} max\{\frac{H(\textit{dfr}(a,L\upharpoonright_{B}))-H(\textit{dfr}(a,L\upharpoonright_{A}))}{H(\textit{dfr}(a,L\upharpoonright_{B}))} ,\\ \frac{H(\textit{dpr}(a,L\upharpoonright_{B}))-H(\textit{dpr}(a,L\upharpoonright_{A}))}{H(\textit{dpr}(a,L\upharpoonright_{B}))}\}.$

We choose a threshold parameter $r$ to indicate the minimum value of MRIG for considering a projection set as potentially interesting. Like in the log entropy based approach, we start from the set of elementary log projection sets, $S_{1}=\{\{a\}|a\in\Sigma_L\}$ and $S'_1=S_1$. We proceed iteratively defining $S_{i+1}=\{A\cup B|A\in S'_{i},B\in S_{1}:B\not\subseteq A\}$ and $S_{i+1}'=\{A|A\in S_{i+1} \wedge \exists B\in S_{i}':\mathit{MRIG}(A,B,L)>r\}$, filtering out those projection sets where adding an extra activity to the projection set leads to an insufficient decrease in the number bits of entropy needed to encode the log statistics. 
The procedure stops if $S'_{i+1}=\emptyset$ or $S_{i+1}=\Sigma_L$. Again, projection sets that are subsets of other projection sets are finally removed: $S_{\textit{final}} = \{A|\exists S'_i: A\in S'_i \wedge (\forall S'_j ,\forall B\in S'_j: A\not\subset B)\}$.

\section{Experimental setup}
\label{sec:evaluation_methodology}
\begin{figure}
	\centering
	\includegraphics[width=0.49\textwidth]{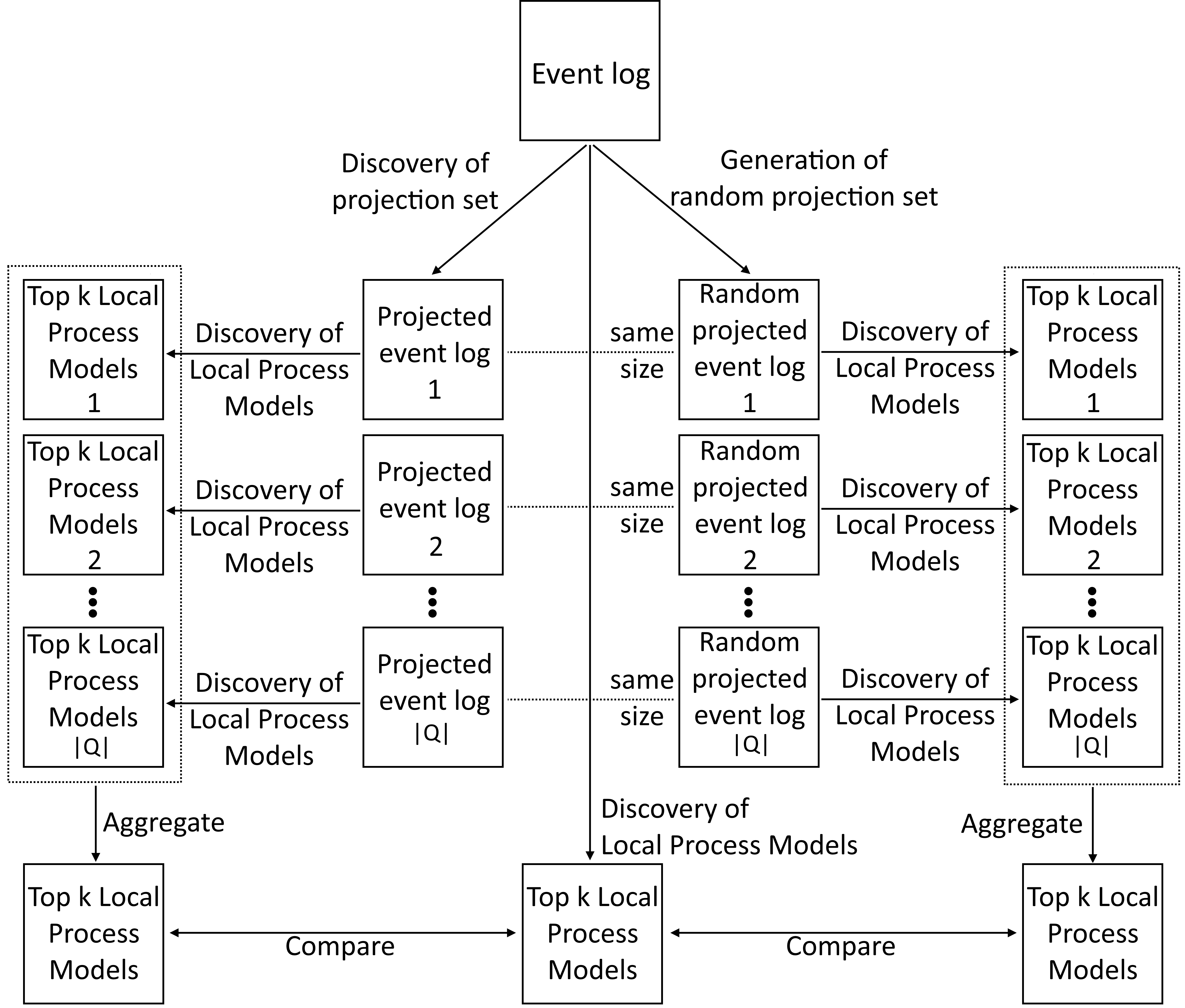}
	\caption{Evaluation methodology for projection set discovery approaches}
	\label{fig:experimental_setup}
	\vspace{-0.2cm}
\end{figure}

Fig. \ref{fig:experimental_setup} gives an overview of the methodology for evaluation of projection set discovery methods. We assess their quality via the comparison of the quality of the Local Process Models (LPMs) discovered using the projection sets that they generate with the quality of the LPMs discovered on the same log without the use of projections. First, we apply LPM discovery to the original, unprojected, event log $L$, resulting in the ``ideal'' top $k$ of LPMs. Then we apply one of the projection set discovery methods on event log $L$, resulting in a set $Q$ of projection sets. On each of the projected event logs in $L\tproj_Q$ we apply LPM Discovery, resulting in $|Q|\times k$ LPMs. We select from them $k$ unique best LPMs in terms of weighted average over the quality criteria. We compare the ``ideal'' set of LPMs with the set discovered using projections. Theoretically, they can coincide, or be equally good, if for every ``ideal'' LPM there is a projection set containing its activities. If it is not the case, some of the best scoring models will be missing and we compare how close are the scores of the models present in the set to the scores of the ``ideal'' LPMs.

Even with the loss of quality in LPM, projections could be a good option to opt for, since they allow to significantly improve the time performance and make LPM discovery possible for logs with many different activities. We evaluate how our projection discovery methods perform compared to random projections as follows: We create a set $Q'$ of random projection sets, consisting of $|Q|$ projection sets where for each projection set $q\in Q$ we create a random projection set $q'\subseteq \Sigma_L$ of the same size as $q$: $|q'|=|q|$. We apply LPM discovery on each of the projected event logs in $L\tproj_{Q'}$ and select $k$ best LPMs from the discovered ones. The projection discovery method works well if the LPMs from top $k$ score closer to the ``ideal'' top $k$ than the LPMs generated based on the random projection sets. To obtain statistically relevant results, we create the random projection sets ten times.

\subsection{Metrics for Comparison of Local Process Model Rankings}
We define the recall, denoted \emph{recall@k}, as the fraction of LPMs from the ``ideal'' top $k$ that are also discovered with the use of projections. A more nuanced way of comparison is to look at the weighted average scores of the LPMs in the ``ideal'' ranking. The intuition behind this is that not being able to find LPMs from the ``ideal'' top $k$ is less severe in case the alternatively found LPMs are also of good quality, i.e. just below the top $k$. Furthermore, missing the best-scoring LPMs is more severe than missing the lower scoring LPMs. For this reason, we use Normalized Discounted Cumulative Gain (NDCG@k) \cite{Jarvelin2002,Burges2005}, one of the most widely used metrics for evaluation of a ranking with an ``ideal'' ranking in the field of Information Retrieval \cite{Tax2015}, that gives more weight to the top of the rankings than to the lower parts of the rankings.

Discounted Cumulative Gain (DCG) aggregates the relevant scores of the individual LPMs in the ranking of LPMs in such a way that the graded relevance is reduced in the logarithmic proportion to the position of the result; due to that, more weight is put on the top of the ranking than on the lower parts of the ranking. DCG is formally defined as: $\mathit{DCG@k} = \sum_{i=1}^k \frac{2^{rel_i}-1}{log_2(i+1)}$,\looseness=-1

\noindent where $rel_i$ is defined as the relevance, which is the weighted average of the model's scores on the quality dimensions for the LPM on position $i$ in the ranking. Normalized Discounted Cumulative Gain (NDCG) is obtained by dividing the DCG value by the theoretical maximum of the discounted cumulative gain value, which is called Ideal Discounted Cumulative Gain (IDCG). The IDCG value is the DCG value obtained from the ground truth ranking on the LPMs discovered on the original, non-projected log, since all local process models that can be discovered from projected event logs can also be discovered from the original event log. Normalized Discounted Cumulative Gain (NDCG) is defined as:

$\mathit{NDCG@k} = \frac{\mathit{DCG@k}}{\mathit{IDCG@k}}$.

\subsection{Data Sets for Projection Discovery Experiments}
We perform the evaluation on five different data sets using the methodology described above. All of the data sets used originate from the human behavior logging domain, with four data sets consisting of \emph{Activities of Daily Life} (ADL) and one consisting of activities performed by an employee in a working environment. Event logs from the human behavior domain are generally too unstructured to allow for discovery of informative process models with process discovery techniques, while LPM discovery allows to discover some relations between activities at a local level that are not discoverable with usual process mining techniques. Table \ref{tab:data_sets} gives an overview of main event log characteristics used in the evaluation. The event logs used for in the experiments have limited number of activities (at most 14), which allows us to determine the ground truth ranking of LPMs, that is, the ranking of LPMs obtained by LPM discovery on the full log without using projections. However, log projections also enable discovery of LPMs on datasets with many more activities, such as the LPMs in Fig. \ref{fig:motivating_example} that were discovered on a log with 1734 activities. For the Van Kasteren data set \cite{Kasteren2008} we show and discuss the ranking of LPMs for which we aim to speed up discovery through projections. For each data set we identified a support threshold used for pruning that allows us to run LPM discovery on the unprojected event log within reasonable time (max. 10 minutes on a 4-core 2.4 GHz CPU). This value depends on the number of activities as well as the length of the traces within the event logs, i.e. more activities and/or longer traces result in a need for a higher support threshold to finish the experiment within the time limit.\looseness=-1

\begin{table}
	\centering
	\caption{An overview of the data sets used in the evaluation experiment}
	\label{tab:data_sets}
	\scalebox{0.7}{
		\begin{tabular}{|l|r|r|r|}
			\toprule
			Data set & \# of activities & \# of cases & \# of events \\
			\midrule
			BPI '12 resource 10939& 14 & 49 & 1682\\
			Bruno & 14 & 57 & 553 \\
			CHAD subject 1900010 & 10 & 26 & 238\\
			Ordonez A & 12 & 15 & 409\\
			Van Kasteren & 14 & 23 & 1285\\
			\bottomrule
		\end{tabular}}
		\vspace{-0.3cm}
	\end{table}

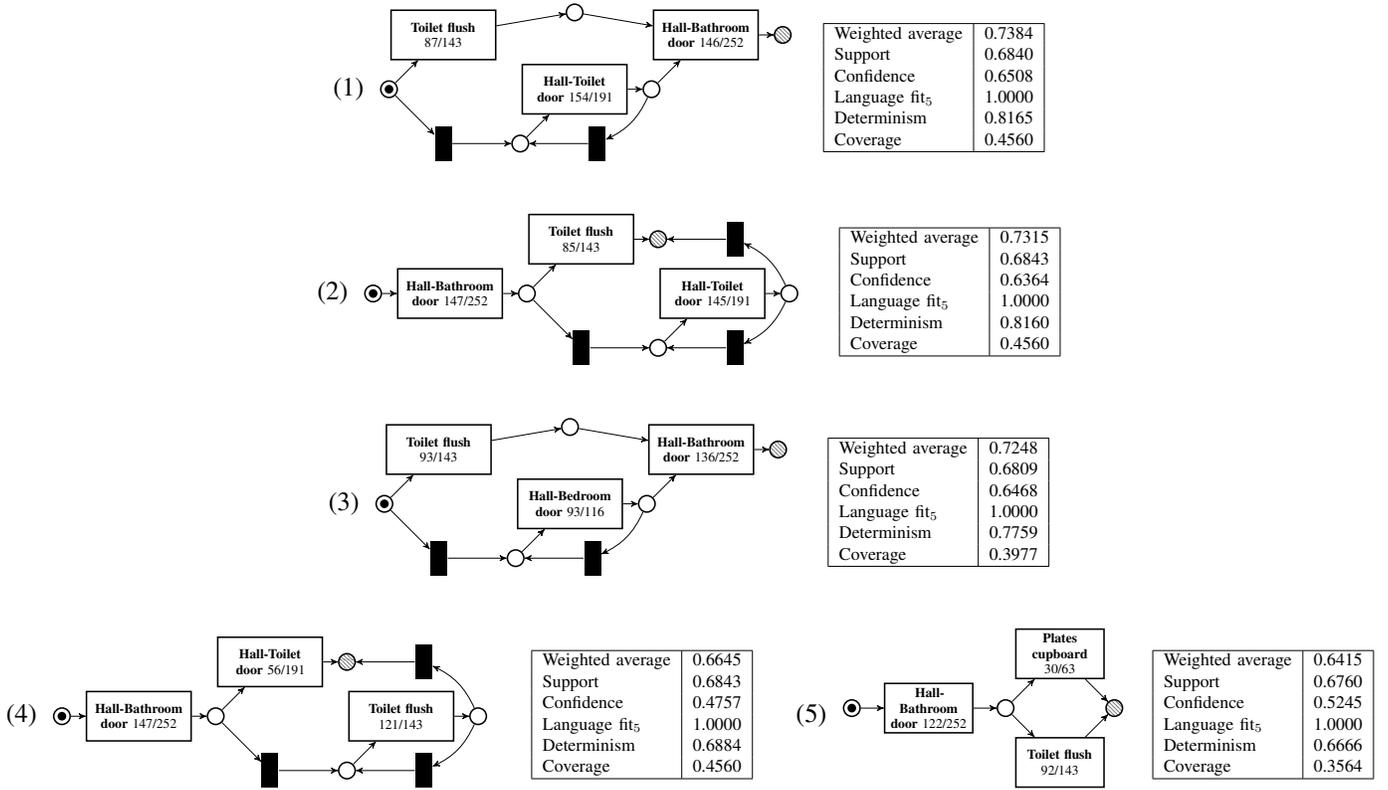
\begin{figure*}[t]
\captionsetup[subfigure]{labelformat=empty}
\centering
\subfloat{
	\raggedleft
	\subfloat{
	\raggedleft
	\raisebox{3.65\height}{(1)}
	\scalebox{0.73}{
		\begin{tikzpicture}
		[node distance=1.4cm,
		on grid,>=stealth',
		bend angle=20,
		auto,
		every place/.style= {minimum size=1mm},
		every transition/.style = {minimum size = 7mm, text width=1.8cm,align=center},
		transitionH/.style={rectangle, thick, fill=black, minimum width=3mm, inner ysep=9pt }
		]
		\node [place, tokens = 1] (p){};
		\node [transition] (t1) [above right = of p,font=\scriptsize]{\textbf{Toilet flush} 87/143}
		edge[pre] node[auto] {} (p);
		\node [transitionH] (t3) [below right = of p,fill=black] {}
		edge[pre] node[auto] {} (p);
		\node [place] (p8) [right=of t3]{}
		edge[pre] node[auto] {}(t3);
		\node [transitionH] (t4) [right = of p8,fill=black]{}
		edge[post] node[auto] {}(p8);
		\node [place] (p9) [above right = of t4]{}
		edge[post, bend left] node[auto] {}(t4);
		\node [transition] (t6) [left = of p9,font=\scriptsize]{\textbf{Hall-Toilet door} 154/191}
		edge[post] node[auto] {}(p9)
		edge[pre] node[auto] {}(p8);
		\node [transition] (2) [above right = of p9,font=\scriptsize] {\textbf{Hall-Bathroom door} 146/252}
		edge [pre] node[auto] {} (p9);
		\node [place,pattern=custom north west lines,hatchspread=1.5pt,hatchthickness=0.25pt,hatchcolor=gray] (p3) [right = of 2] {}
		edge[pre] node[auto] {} (2);
		\node [place] (p2) [above = of t6] {}
		edge[pre] node[auto] {} (t1)
		edge[post] node[auto] {} (2);
		\end{tikzpicture}
	}}
	\subfloat{
		\raggedleft
		\raisebox{\height}{
			\scalebox{0.67}{
				\begin{tabular}{|l|l|}
					\hline
					Weighted average & 0.7384\\
					Support & 0.6840\\
					Confidence& 0.6508\\
					Language fit$_5$& 1.0000\\
					Determinism& 0.8165\\
					Coverage& 0.4560\\
					\hline
				\end{tabular}
			}
		}
	}
}
\\

\subfloat{
	\raggedleft
	\subfloat{
		\raggedleft
		\raisebox{3.65\height}{(2)}
		\scalebox{0.73}{
			\begin{tikzpicture}
			[node distance=1.4cm,
			on grid,>=stealth',
			bend angle=20,
			auto,
			every place/.style= {minimum size=1mm},
			every transition/.style = {minimum size = 7mm, text width=1.8cm,align=center},
			transitionH/.style={rectangle, thick, fill=black, minimum width=3mm, inner ysep=9pt }
			]
			\node [place, tokens = 1] (p){};
			\node [transition] (2) [right = of p,font=\scriptsize] {\textbf{Hall-Bathroom door} 147/252}
			edge [pre] node[auto] {} (p);
			\node [place] (p3) [right = of 2] {}
			edge[pre] node[auto] {} (2);
			\node [transition] (t1) [above right = of p3,font=\scriptsize]{\textbf{Toilet flush} 85/143}
			edge[pre] node[auto] {} (p3);
			\node [transitionH] (t3) [below right = of p3,fill=black] {}
			edge[pre] node[auto] {} (p3);
			\node [place] (p8) [right=of t3]{}
			edge[pre] node[auto] {}(t3);
			\node [transitionH] (t4) [right = of p8,fill=black]{}
			edge[post] node[auto] {}(p8);
			\node [place] (p9) [above right = of t4]{}
			edge[post, bend left] node[auto] {}(t4);
			\node [transition] (t6) [left = of p9,font=\scriptsize]{\textbf{Hall-Toilet door} 145/191}
			edge[post] node[auto] {}(p9)
			edge[pre] node[auto] {}(p8);
			\node [transitionH] (t5) [above left = of p9,fill=black]{}
			edge[pre, bend left] node[auto] {}(p9);
			\node [place,pattern=custom north west lines,hatchspread=1.5pt,hatchthickness=0.25pt,hatchcolor=gray] (p7) [right=of t1] {}
			edge[pre] node[auto] {}(t1)
			edge[pre] node[auto] {}(t5);
			\end{tikzpicture}
		}
	}
	\subfloat{
		\raggedleft
		\raisebox{\height}{
			\scalebox{0.67}{
				\begin{tabular}{|l|l|}
					\hline
					Weighted average & 0.7315\\
					Support & 0.6843\\
					Confidence& 0.6364\\
					Language fit$_5$& 1.0000\\
					Determinism& 0.8160\\
					Coverage& 0.4560\\
					\hline
				\end{tabular}
			}
		}
	}
}
\\

\subfloat{
	\raggedleft
	\subfloat{
		\raggedleft
		\raisebox{3.65\height}{(3)}
		\scalebox{0.73}{
			\begin{tikzpicture}
			[node distance=1.4cm,
			on grid,>=stealth',
			bend angle=20,
			auto,
			every place/.style= {minimum size=1mm},
			every transition/.style = {minimum size = 7mm, text width=1.8cm,align=center},
			transitionH/.style={rectangle, thick, fill=black, minimum width=3mm, inner ysep=9pt }
			]
			\node [place, tokens = 1] (p){};
			\node [transition] (t1) [above right = of p,font=\scriptsize]{\textbf{Toilet flush} 93/143}
			edge[pre] node[auto] {} (p);
			\node [transitionH] (t3) [below right = of p,fill=black] {}
			edge[pre] node[auto] {} (p);
			\node [place] (p8) [right=of t3]{}
			edge[pre] node[auto] {}(t3);
			\node [transitionH] (t4) [right = of p8,fill=black]{}
			edge[post] node[auto] {}(p8);
			\node [place] (p9) [above right = of t4]{}
			edge[post, bend left] node[auto] {}(t4);
			\node [transition] (t6) [left = of p9,font=\scriptsize]{\textbf{Hall-Bedroom door} 93/116}
			edge[post] node[auto] {}(p9)
			edge[pre] node[auto] {}(p8);
			\node [transition] (2) [above right = of p9,font=\scriptsize] {\textbf{Hall-Bathroom door} 136/252}
			edge [pre] node[auto] {} (p9);
			\node [place,pattern=custom north west lines,hatchspread=1.5pt,hatchthickness=0.25pt,hatchcolor=gray] (p3) [right = of 2] {}
			edge[pre] node[auto] {} (2);
			\node [place] (p2) [above = of t6] {}
			edge[pre] node[auto] {} (t1)
			edge[post] node[auto] {} (2);
			\end{tikzpicture}
		}
	}
	\subfloat{
		\raggedleft
		\raisebox{\height}{
			\scalebox{0.67}{
				\begin{tabular}{|l|l|}
					\hline
					Weighted average & 0.7248\\
					Support & 0.6809\\
					Confidence& 0.6468\\
					Language fit$_5$& 1.0000\\
					Determinism& 0.7759\\
					Coverage& 0.3977\\
					\hline
				\end{tabular}
			}
		}
	}
}
\\
			
\subfloat{
	\hspace{-0.42cm}
	\raggedleft
	\subfloat{
		\raggedleft
		\raisebox{3.65\height}{(4)}
		\scalebox{0.73}{
			\begin{tikzpicture}
			[node distance=1.4cm,
			on grid,>=stealth',
			bend angle=20,
			auto,
			every place/.style= {minimum size=1mm},
			every transition/.style = {minimum size = 7mm, text width=1.8cm,align=center},
			transitionH/.style={rectangle, thick, fill=black, minimum width=3mm, inner ysep=9pt }
			]
			\node [place, tokens = 1] (p){};
			\node [transition] (2) [right = of p,font=\scriptsize] {\textbf{Hall-Bathroom door} 147/252}
			edge [pre] node[auto] {} (p);
			\node [place] (p3) [right = of 2] {}
			edge[pre] node[auto] {} (2);
			\node [transition] (t1) [above right = of p3,font=\scriptsize]{\textbf{Hall-Toilet door} 56/191}
			edge[pre] node[auto] {} (p3);
			\node [transitionH] (t3) [below right = of p3,fill=black] {}
			edge[pre] node[auto] {} (p3);
			\node [place] (p8) [right=of t3]{}
			edge[pre] node[auto] {}(t3);
			\node [transitionH] (t4) [right = of p8,fill=black]{}
			edge[post] node[auto] {}(p8);
			\node [place] (p9) [above right = of t4]{}
			edge[post, bend left] node[auto] {}(t4);
			\node [transition] (t6) [left = of p9,font=\scriptsize]{\textbf{Toilet flush} 121/143}
			edge[post] node[auto] {}(p9)
			edge[pre] node[auto] {}(p8);
			\node [transitionH] (t5) [above left = of p9,fill=black]{}
			edge[pre, bend left] node[auto] {}(p9);
			\node [place,pattern=custom north west lines,hatchspread=1.5pt,hatchthickness=0.25pt,hatchcolor=gray] (p7) [right=of t1] {}
			edge[pre] node[auto] {}(t1)
			edge[pre] node[auto] {}(t5);
			\end{tikzpicture}
		}
	}
	\hspace{-0.1cm}
	\subfloat{
		\hspace{-0.1cm}
		\raggedleft
		\raisebox{\height}{
			\scalebox{0.67}{
				\begin{tabular}{|l|l|}
					\hline
					Weighted average & 0.6645\\
					Support & 0.6843\\
					Confidence& 0.4757\\
					Language fit$_5$& 1.0000\\
					Determinism& 0.6884\\
					Coverage& 0.4560\\
					\hline
				\end{tabular}
			}
		}
	}
}
\subfloat{
	\raggedleft
	\subfloat{
		\raggedleft
		\raisebox{3.65\height}{(5)}
		\scalebox{0.73}{
			\begin{tikzpicture}
			[node distance=1.4cm,
			on grid,>=stealth',
			bend angle=20,
			auto,
			every place/.style= {minimum size=1mm},
			every transition/.style = {minimum size = 5mm, text width=1.5cm,align=center}
			]
			\node [place, tokens = 1] (p){};
			\node [transition] (2) [right = of p,font=\scriptsize] {\textbf{Hall-Bathroom door} 122/252}
			edge [pre] node[auto] {} (p);
			\node [place] (p3) [right = of 2] {}
			edge[pre] node[auto] {} (2);
			\node [transition] (t1) [above right = of p3,font=\scriptsize]{\textbf{Plates cupboard} 30/63}
			edge[pre] node[auto] {} (p3);
			\node [transition] (t3) [below right = of p3,font=\scriptsize] {\textbf{Toilet flush} 92/143}
			edge[pre] node[auto] {} (p3);
			\node [place,pattern=custom north west lines,hatchspread=1.5pt,hatchthickness=0.25pt,hatchcolor=gray] (p7) [above right=of t3] {}
			edge[pre] node[auto] {}(t3)
			edge[pre] node[auto] {}(t1);
			\end{tikzpicture}
		}
	}
	\hspace{-0.2cm}
	\subfloat{
		\hspace{-0.2cm}
		\raggedleft
		\raisebox{\height}{
			\scalebox{0.67}{
				\begin{tabular}{|l|l|}
					\hline
					Weighted average & 0.6415\\
					Support & 0.6760\\
					Confidence& 0.5245\\
					Language fit$_5$& 1.0000\\
					Determinism& 0.6666\\
					Coverage& 0.3564\\
					\hline
				\end{tabular}
			}
		}
	}
}
\caption{Example top 5 Local Process Models discovered from the Van Kasteren data set.}
\label{fig:van_kasteren_local_process_models}
\vspace{-0.15cm}
\end{figure*}

\subsubsection{BPIC '12 Resource 10939 Data Set}
The Business Process Intelligence Challenge (BPIC)'12 data set originates from a personal loan application process in a global financial institution. We transformed the event log to obtain traces of daily activities of one specific employee. We set the pruning parameter to $0.675$ for this data set.
\subsubsection{Bruno Data Set}
The Bruno et al. \cite{Bruno2013} data set is a public collection of labeled wrist-worn accelerometer recordings. The data set is composed of fourteen types of ADL events performed by sixteen volunteers. We set the pruning parameter to $0.6$ for this data set.
\subsubsection{CHAD Data Set}
The CHAD database \cite{Mccurdy2000} consists of 22 exposure and time-use studies that have been consolidated in a consistent format. In total the database contains 54000 individual days of human behavior, from which we extract an event log from a randomly chosen study subject such that each case represents a day. For this data set we set the support pruning parameter to $0.4$.
\subsubsection{Ordonez A Data Set}
The Ordonez \cite{Ordonez2013} data set consists of ADL events that are performed by two users in their own homes, recorded through smart home sensors. We use an event log obtained from sensor events of subject A, with each case representing a day. We set the pruning parameter to $0.6$.
\subsubsection{Van Kasteren Data Set}
This data set is a smart home environment event log described by Van Kasteren et al. \cite{Kasteren2008}. It consists of multidimensional time series data, where each dimension represents the binary state of an in-home sensor. These sensors include motion sensors, open/close sensors, and power sensors (discretized to on/off states). We transform the multidimensional sensor time series into events by considering the change points of sensor states as events. We create cases by grouping events by day, with a cut-off point at midnight. We set the pruning parameter to $0.675$ for this data set. 

Fig. \ref{fig:van_kasteren_local_process_models} shows the first five elements of the Local Process Model ranking discovered on one of the data sets (Van Kasteren \cite{Kasteren2008}). The $5^{\textit{th}}$ LPM shows that roughly half of the times that the \emph{Hall-Bathroom door} is opened, either \emph{Toilet flush} or \emph{Plates cupboard} are observed afterwards. The bathroom in this house contains both a toilet and a shower, but showering events are not observed as the shower does not contain a sensor. It is likely that \emph{Hall-Bathroom} events that are followed by \emph{Plates cupboard} events represent a morning shower after which the subject proceeds his morning routine with breakfast. The ordering of LPMs matches the weighted average of the LPM quality criteria.\looseness=-1

\section{Results \& Discussion}
\label{sec:results}
\begin{figure*}
	\centering
	\includegraphics[width=0.74\textwidth]{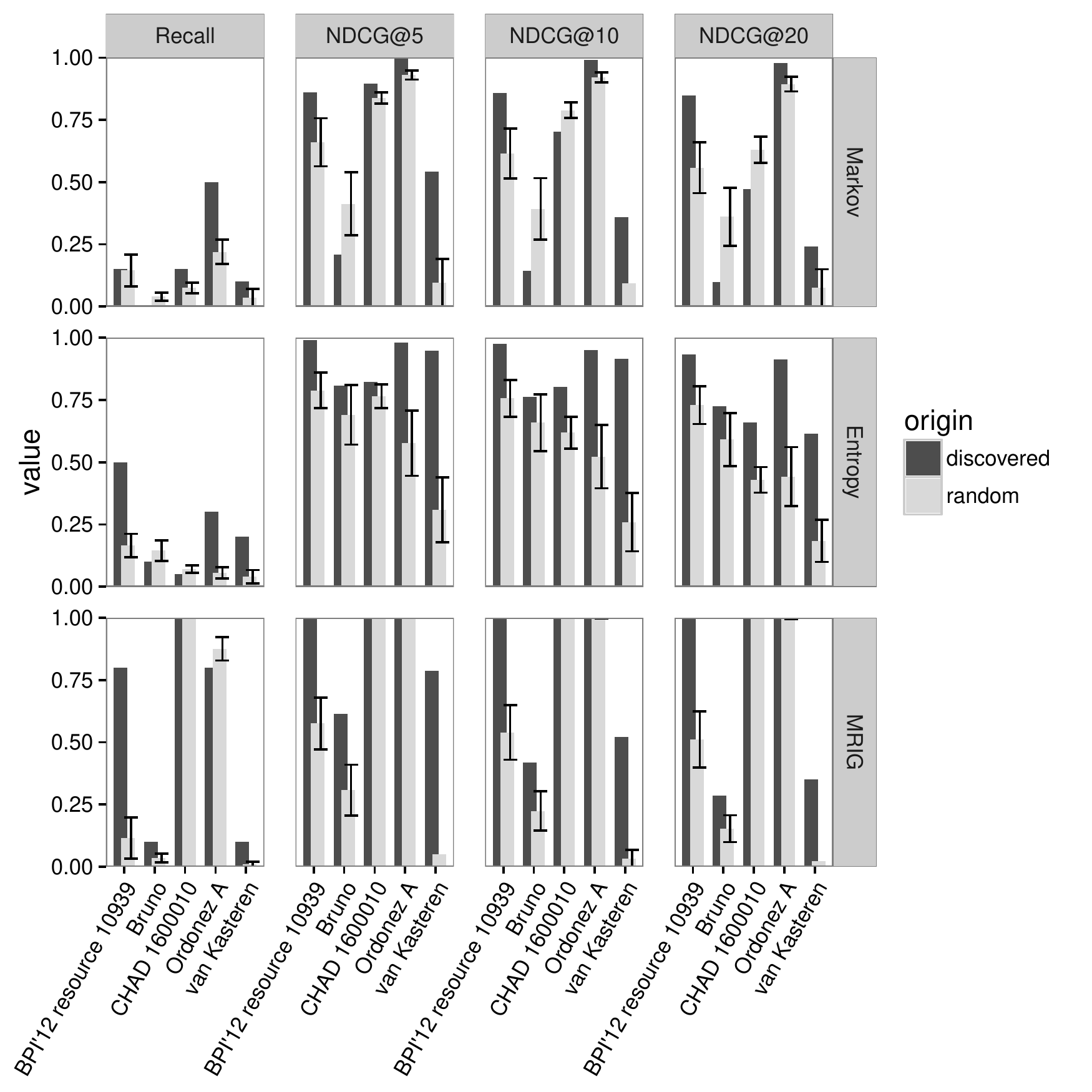}
	\caption{Performance of the three projection set discovery methods on the six data sets on the four metrics}
	\label{fig:results}
	\vspace{-0.2cm}
\end{figure*}

Fig. \ref{fig:results} shows the results of the evaluation using the five evaluation data sets and Table \ref{tab:speedup_results} shows the speedup of projection-based LPM discovery. The speedup mainly depends on the size of the projection sets. Note that we do not compare the computation time of the discovered and the randomly generated projections, as both consist of equally sized projections. The time needed for discovering the projection set is included in the computation time shown in Table \ref{tab:speedup_results}, however, the projection set discovery time is negligible compared to the time needed for LPM discovery. Each dark gray bar in Fig. \ref{fig:results} represents the performance on the measure indicated by the column of the projection set discovery method indicated by the row on the dataset indicated below. Each gray bar indicates the average performance over ten random projection sets of the same size as the discovered projection set belonging to the dark gray bar to its left. The error bars indicate the Standard Error (SE), indicating the uncertainty of the mean performance of the random projection sets.\looseness=-1

Our first observation is that all three projection set discovery methods are better than random projections on almost all data sets. The only exception is the Markov-based projection set discovery approach in case of the Bruno data set, which performs worse than random projections, with none of the LPMs discovered on the original data set being found on the projection sets. The Markov clustering based approach to projection set discovery achieves the highest speedup on three of the five data sets, and second highest speedup on the other two data sets. The high speedup indicates that the projections generated by Markov clustering are relatively small, which also explains the quality loss of the LPMs with regard to the ground truth which is typically larger than with the other projection set discovery methods. The size of the clusters created with Markov clustering can be influenced through its inflation parameter, which we set to $1.5$. We found lower values to result in all activities being clustered into one single cluster, meaning that the only projection created is the original log itself.\looseness=-1

The entropy based approach shows a higher gain than the Markov based approach on the NDCG@\{5,10,20\} metrics for all data set, when compared to random projections of the same size. However, the obtained speedup of LPM discovery with the entropy based approach is lower than speedup with the Markov based approach on all but one data set.

The MRIG based approach shows significant improvements on all metrics on the BPI'12 and Bruno data sets. On these two data sets it also results in the second highest and highest speedup respectively. However, on the CHAD data and the Ordonez data the discovered projection sets consist of a single projection that contains almost all log activities, resulting consequently in no speedup with close to perfect recall scores.

\begin{table}
\centering
	\caption{Local Process Model discovery speedup obtained with each projection set discovery method on the evaluation data sets\looseness=-1}
	\centering
	\scalebox{0.69}{
	\begin{tabular}{|l|r|r|r|}
		\toprule
		Data set & Speedup Markov & Speedup entropy & Speedup MRIG \\
		\midrule
		BPI'12 resource 10939 & \textbf{42.9} & 5.6 & 6.9 \\
		Bruno & 222.9 & 111.5 & \textbf{336.1} \\
		CHAD subjection 1900010 & \textbf{1060.1} & 1041.2 & 1 \\
		Ordonez A & 7.3 & \textbf{33.7} & 1.1 \\
		Van Kasteren & \textbf{111.2} & 83.5 & 96.7 \\
		\bottomrule
	\end{tabular}}
	\label{tab:speedup_results}
	\vspace{-0.4cm}
\end{table}

\section{Related Work}
\label{sec:related}
The task of discovering projections plays an important role within the area of decomposed process discovery and conformance checking. Decomposed process discovery aims at partitioning the activities in the event log such that after applying process discovery to each partition of the events, the start-to-end process model can be constructed by stitching together the process models discovered on the individual partitions. In \cite{Aalst2013} an approach was introduced to decompose process mining by using a \emph{maximal decomposition} of a causal dependency graph, where the activities associated with each edge in the causal dependency graph end up in one cluster. Hompes et al. \cite{Hompes2014} describe an approach to make more coarse-grained activity clusters by recombining the clusters of the maximal decomposition by balancing three quality criteria: cohesion, coupling, and balance. Van der Aalst and Verbeek \cite{Aalst2014} introduced a decomposed process mining approach based on \emph{passages}. A passage is a pair of two non-empty sets of activities (X,Y) such that the set of direct successors of X is Y and the set of direct predecessors of Y is X. Munoz-Gama et al. \cite{Munoz2014} proposed a decomposed conformance checking approach that discovers clusters of activities based on identifying Single-Entry Single-Exit (SESE) blocks in a Petri net model of the process. A SESE block in a Petri net is a set of edges that has exactly two boundary nodes: one entry and one exit. Clustering activities using this approach assumes availability of a structured process model describing process instances from the beginning to the end, which is different from the application area of LPM discovery.\looseness=-1

Carmona et al. \cite{Carmona2009} describe an approach to generate overlapping sets of activities from a causal dependency graph and uses it to speed up region theory based Petri net synthesis from a transitions system representation of the event log. The activities are grouped together in such a way that the synthesized Petri nets can be combined into a single Petri net generating all the traces of the log. In a more recent paper Carmona \cite{Carmona2012} describes an approach to discover a set of projections from an event log based on Principle Component Analysis (PCA).\looseness=-1

All the projection discovery methods mentioned above aim at the discovery of models that can be combined into a single process model. In the context of LPM discovery, we are not interested in combining process models into one single process model; instead, each LPM is assumed to convey interesting information about a relationship between activities itself. Projection methods in the context of decomposed process mining all aim to minimize overlap between projections, leaving only the overlap needed for combining the individual process models into one. In our case, overlap between clusters is often desired, as interesting patterns might exist within a set of activities $\{A,B,C,D\}$, as well as within a set of activities $\{A,B,C,E\}$, and discovering on both subsets individually is faster than discovering once on $\{A,B,C,D,E\}$. The three projection set discovery methods introduced in this paper exploit this and aim for overlapping projection sets.

The episode miner \cite{Leemans2014} is a related technique to LPM discovery, which discovers patterns that are less expressive (i.e. limited to partial orders), but is computationally less expensive. While the need for heuristic techniques to speed up episode mining is limited compared to LPM discovery, in principle the three heuristics described to speedup LPM discovery could also be used to speedup the discovery of frequent episodes

\section{Conclusion \& Future Work}
We explored three different heuristics for the discovery of projection sets for speeding up Local Process Model (LPM) discovery. These heuristics enable the discovery of LPMs from event logs where it is computationally not feasible to discover LPMs from the full set of activities in the log. All three of them produce better than random projections on a variety of data sets. Projection discovery based on Markov clustering leads to the highest speedup, while higher quality LPMs can be discovered using a projection discovery based on log statistics entropy. The Maximal Relative Information Gain based approach to projection discovery shows unstable performance with the highest gain in LPM quality over random projections on some event logs, while not being able to discover any projection smaller than the complete set of activities on some other event logs. In fact, we would like to explore event log properties that can serve as a predictor for the relative performance of these methods.\looseness=-1
\label{sec:conclusion}



%

\bibliographystyle{ieeetran}
\bibliography{arxiv}

\end{document}